\documentclass[runningheads]{llncs}
\usepackage{wrapfig}
\usepackage{subfig}
\usepackage{amsmath}
\usepackage{graphicx}
\usepackage{algorithm2e}
\usepackage{algorithmic}	
\usepackage{amsfonts}
\usepackage{pgfplots}
\usepackage{pgfplotstable}
\usetikzlibrary{plotmarks}
\usepackage{pgfgantt}
\newcommand{\plotred}{\addplot[color=red, mark=ball]}

\newcommand{\plotblue}{\addplot[color=blue, mark=otimes]}
\newcommand{\plotblack}{\addplot[color=black, mark=star]}
\newcommand{\plotyellow}{\addplot[color=yellow!60!red, mark=diamond] }

\def\addlegendimage{\csname pgfplots@addlegendimage\endcsname}
\newenvironment{customlegend}[1][]{%
	\begingroup
	\csname pgfplots@init@cleared@structures\endcsname
	\pgfplotsset{#1}%
}{%
	\csname pgfplots@createlegend\endcsname
	\endgroup
}%
\pgfplotsset{compat=1.13}
\definecolor{Gray}{gray}{0.95}
\definecolor{LGray}{gray}{0.75}
\definecolor{bblue}{HTML}{4F81BD}
\definecolor{rred}{HTML}{C0504D}
\definecolor{ggreen}{HTML}{9BBB59}
\definecolor{ppurple}{HTML}{9F4C7C}
\usetikzlibrary{shapes.geometric,arrows.meta} 
\tikzset{vertex style/.style={
		draw=#1,
		thick,
		fill=#1!70,
		text=white,
		ellipse,
		minimum width=2cm,
		minimum height=0.75cm,
		font=\small,
		outer sep=0.3pt,
	},
	text style/.style={
		sloped,
		text=black,
		font=\footnotesize,
		above
	}
}
\tikzset{
	vector/.pic={
		\draw (0,0) rectangle (0.5,0.25) rectangle (1,0) rectangle (1.5,0.25) rectangle (2,0); 
	}
}

\usepackage{color}
\newcommand{\system}{\textsf{Triple2Vec}}
\newcommand{\kgs}{{KGs}}
\newcommand{\kg}{{G}}

\newcommand{\lkg}{\mathcal{G}_{L}}

\newtheorem{exmp}[theorem]{\normalfont \textbf{Example}}
\newtheorem{defn}[theorem]{\normalfont  \textbf{Definition}}

\newcommand{\entity}[1]{{\small\textsf{#1}}}

\begin{document}
\title{\system: Learning Triple Embeddings from Knowledge Graphs}
\author{Valeria FIonda\inst{1} \and Giuseppe Pirr\'o\inst{2}}
\authorrunning{Fionda and Pirr\'o.}
\institute{Department of Mathematics and Computer Science, University of Calabria, Italy \and Department of Computer Science, University of Rome La Sapienza, Italy}
%\email{fionda@mat.unical.it, pirro@di.uniroma1.it}

\maketitle              
\begin{abstract}
Graph embedding techniques allow to learn high-quality feature vectors from graph structures and are useful in a variety of tasks, from node classification to clustering. Existing approaches have only focused on learning feature vectors for the nodes in a (knowledge) graph. To the best of our knowledge, none of them has tackled the problem of embedding of graph edges, that is, knowledge graph triples. The approaches that are closer to this task have focused on homogeneous graphs involving only one type of edge and obtain edge embeddings by applying some operation (e.g., average) on the embeddings of the endpoint nodes. The goal of this paper is to introduce \system, a new technique to directly embed edges in (knowledge) graphs. \system\ builds upon three main ingredients. \textit{The first} is the notion of line graph. The line graph of a graph is another graph representing the adjacency between edges of the original graph. In particular, the nodes of the line graph are the edges of the original graph. We show that directly applying existing embedding techniques on the nodes of the line graph to learn edge embeddings is not enough in the context of knowledge graphs. Thus, we introduce the notion of triple line graph. \textit{The second} is an edge weighting mechanism both for line graphs derived from knowledge graphs and homogeneous graphs. \textit{The third }is a strategy based on graph walks on the weighted triple line graph that can preserve proximity between nodes. Embeddings are finally generated by adopting the SkipGram model, where sentences are replaced with graph walks. We evaluate our approach on different real world (knowledge) graphs and compared it with related work.  
%%%%%%%%
\end{abstract}
\section{Introduction}
\label{sec:introduction}
 In the last years, learning graph representations using low-dimensional vectors has received attention as viable support to various (machine) learning tasks, from node classification to clustering \cite{cai2018comprehensive}. Approaches like DeepWalk \cite{perozzi2014deepwalk}, node2vec \cite{grover2016node2vec} and their variants strive to find node representations that preserve structural relations in the learned space. These approaches only focus on \textit{homogeneous networks}, that is, networks (e.g., social networks) including only one type of edge. Another strand of research focused on embedding nodes in knowledge graphs (aka heterogeneous information networks) characterized by several distinct types of nodes and edges  \cite{wang2017knowledge}. Notable approaches are rdf2vec \cite{ristoski2016rdf2vec}, metapath2vec \cite{dong2017metapath2vec}, and JUST \cite{hussein2018meta}.  One common denominator of both homogeneous and knowledge graph based embedding approaches is the usage of language model techniques. The idea is to consider sequences of nodes in a graph (i.e., random walks) as analogous to sentences in a document; then, the node sequences are fed into models like Skip-gram \cite{mikolov2013distributed} to learn the final node embeddings. Despite the variety of available embedding techniques, to the best of our knowledge, \textit{there is no technique that focuses on the embedding of triples (edges) of (knowledge) graphs}. The closest attempt we are aware of has been done by node2vec, where edge embeddings are learned by applying some operators (e.g., average) to the embeddings of the nodes at edge endpoints. This is insufficient in our context for several reasons. First, this approach has only considered homogeneous graphs; it is not clear how to behave when nodes are linked by more than an edge as in the case of knowledge graphs. Second, it is sub-optimal as it does not directly learn edge embeddings. Third, it does not embed labeled edges (i.e., triples).

\begin{figure}[!t]
	\includegraphics[width=\textwidth]{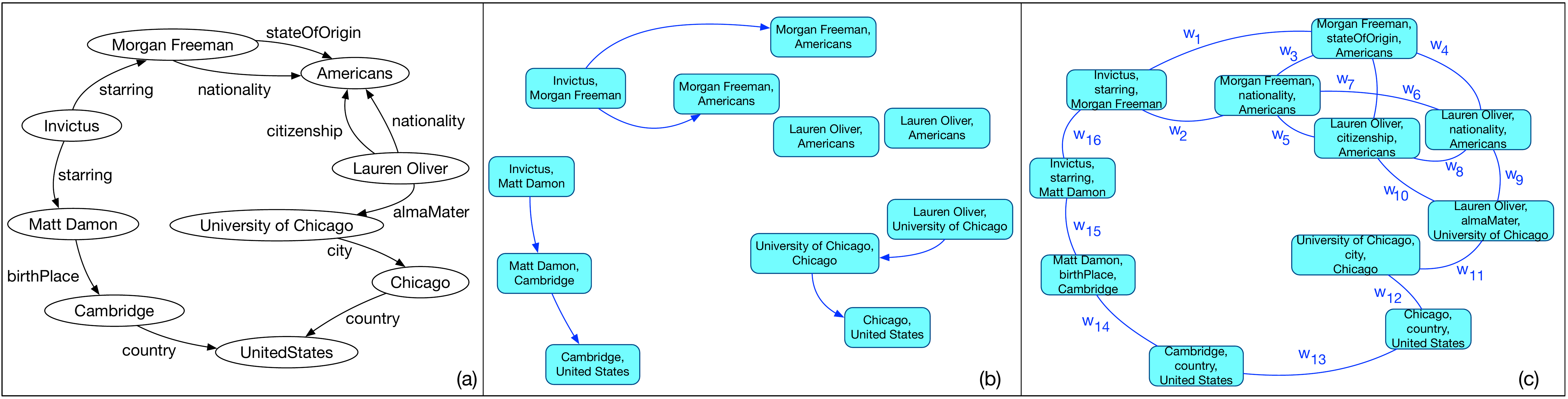}
	\caption{A knowledge graph (a), its directed line graph and (c) its triple line graph.}
	\label{fig:intro_example}
\end{figure}

\textit{The goal of this paper is to devise novel techniques to directly learn edge embeddings from both homogeneous and knowledge graphs.} This sets three main challenges. \textit{The first challenge is about how to go from node embeddings to edge embeddings}. One way to approach the problem could be to perform some (algebraic) manipulation on the endpoints of an edge; nevertheless, it is not clear how to behave in the context of knowledge graphs, where nodes may have multiple edges (i.e., predicates that reflect different relation perspectives) interlinking them. As an example, in Fig. \ref{fig:intro_example} (a) \entity{Lauren Oliver} and \entity{Americans} are linked by two different edge types. To tackle this first challenge, we build upon the notion of \textit{line graph} of a graph \cite{west1996introduction}. In its basic definition, the line graph has as nodes the edges of the original graph (nodes of the line graph are identified by the corresponding edge endpoints) while an edge is added between nodes if they share a common endpoint. This notion has been extended to both directed and multigraphs. 
As an example, the directed line graph obtained from the knowledge graph in Fig. \ref{fig:intro_example} (a) is shown in Fig. \ref{fig:intro_example} (b), where the two copies of the node \entity{Lauren Oliver}, \entity{Americans} correspond to the triples having \entity{nationality} and \entity{citizenship} as a predicate, respectively. It would be tempting to directly apply embedding techniques to the nodes of the directed line graph to obtain edge embeddings. However, we detect two main problems. The first is that it is impossible to discern between the two triples encoded by the nodes  \entity{Lauren Oliver} and \entity{Americans}. The second is that the directed line graph is disconnected and as such, it becomes problematic to learn triple embeddings via random walks. Therefore, we introduce the notion of \textit{triple line graph} $\lkg$ of a knowledge graph $\kg$; here, nodes are the triples of $\kg$ and \textit{an edge is introduced between nodes in $\lkg$ whenever the triples form $\kg$ they represent share an endpoint}. This guarantees that $\lkg$ is connected if $\kg$ is connected. The triple line graph for the graph in Fig. \ref{fig:intro_example} (a) is shown in Fig. Fig. \ref{fig:intro_example} (c). 

However, the $\lkg$ alone is still not enough; in general, $\lkg$ has a much denser structure than $\kg$. \textit{This introduces the second challenge related to the fact that high degree nodes in $\kg$  get over-represented, in terms of the number of edges, in $\lkg$}. To tackle this seconds challenge, we introduce two mechanisms to weight the edges of $\lkg$. The first, specific for \textit{knowledge graphs,} assigns weights on the basis of predicate relatedness \cite{pirrobuilding}. The weight of an edge between nodes of $\lkg$ is equal to the semantic relatedness between the predicates in the triples of $\kg$ represented by the two nodes. As an example, in Fig. \ref{fig:intro_example} (c) the weight of the nodes of $\lkg$ (\entity{M. Damon, birthPlace, Cambridge}) and (\entity{Cambridge, country,  United States}) will be equal to the relatedness between \entity{birthPlace} and \entity{country}.
The second mechanism, specific for \textit{homogeneous graphs}, leverages the centrality of nodes in the original graph $\kg$. The weight of an edge between the nodes of $\lkg$ $(u,i)$ and $(i,j)$ is computed as a function of the flow centrality of $u$, $i$, and $j$ in $\kg$.

\textit{The third challenge consists of how to compute the edge embeddings from the weighted $\lkg$}. To this hand, we generate truncated random walks, in the form of sequences of nodes, on the weighted triple line graph. Note that weights based on semantic relatedness, for knowledge graphs, will bias the random walker to obtain similar contexts for nodes in the weighted triple line graph linked by related predicates. The underlying idea is that similar context will lead to similar embeddings. Finally, the walks are fed into the Skip-gram model \cite{mikolov2013distributed}, which will give the embeddings of the nodes of the weighted triple line graph that correspond to the embeddings of the original edges (for homogeneous graphs) or triples (for knowledge graphs).
%%%%%%%%%%%%%%
We assembled all the ingredients in \system, which, to the best of our knowledge is the first approach that focuses on the problem of learning \textit{edge embeddings from (knowledge) graphs}. We believe that edge embeddings can open up a novel class of downstream applications that go beyond that based on node embeddings. We are going to discuss applications in the area of edge classification and clustering.

\noindent
\textbf{Contributions and Outline.}
We tackle the problem of learning edge embeddings from (knowledge) graphs. To the best of our knowledge, this is the first work in this direction. We make the following main contributions:
\begin{itemize}
		\itemsep0em
	\item We introduce the notion of triple line graph that extends the notion of line graph to knowledge graphs.
	\item We introduce two different weighting mechanisms for triple line graphs. The first, suitable for knowledge graphs, assigns weights based on the relatedness between predicates in triples. The second one, suitable for homogeneous graphs, weights the edges of the triple line graph based on the centrality of the nodes of the original graph.
	\item We introduce for the first time the notion of triple embedding, that is, a technique that can learn embeddings from triples in knowledge graphs.
	\item We describe novel application scenarios and compare our approach with related work.
\end{itemize}

The remainder of the paper is organized as follows. We introduce some preliminary definitions in Section \ref{sec:preliminaries}. In Section \ref{sec:tripleLineGraph}, we introduce the notion of triple line graph of a knowledge graph along with an algorithm to compute it. Section \ref{sec:triple-embeddings} describes the \system\ approach to learn triple embeddings from both homogeneous and knowledge graphs. In Section \ref{sec:experiments} we discuss an experimental evaluation. Related work is dealt with in Section \ref{sec:related-work}. We draw some conclusions and sketch future work in Section \ref{sec:conclusions}
%%%%%%%%%%%%%
\section{Preliminaries}
\label{sec:preliminaries}

A Knowledge Graph (\kg) is a kind of heterogeneous information network. It is a node and edge labeled directed multigraph $G$=$(V_G,E_G,T_G)$ where $V$ is a set of uniquely identified vertices representing entities (e.g., D. Lynch), $E_G$ a set of predicates (e.g., director) and $T$ a set of triples of the form $(s, p, o)$ representing directed labeled edges, where $s, o  \in V$ and $p \in E_G$. Homogeneous graphs are represented as $G$=$(V_G,E_G)$, where $E_G$ is a set of edges of the form $(i,j)$. In what follows we will use the term edge to refer to both triples (for knowledge graphs) and edges for homogeneous graphs when it is clear from the context.
%%%
%%%
%%%
\subsection{The Line Graph of a Graph}
\label{sec:preliminary-line-graph}
We introduce the  notion of line graph of a graph  starting with the case of  undirected graphs.
The idea of the line graph $\lkg$ of an undirected graph $\kg$ is to represent adjacency information between the edges of $G$. More formally:
\begin{defn}
	Given an undirected graph $\kg=(V_G,E_G)$, where $V_G$ (resp., $E_G$) is the sets of node (resp., edges), its line graph $\lkg=(V_L,E_L)$ is such that: (i) each node of $\lkg$ represents an edge of $\kg$; (ii) two vertices of $\lkg$ are adjacent if, and only if, their corresponding edges in $\kg$ have a node in common.
\end{defn}

Starting from $\kg=(V_G,E_G)$ it is possible to compute the number of nodes and edges of $\lkg=(V_L,E_L)$ as follows: \textit{(i)} the number of nodes of $\lkg$ is equals to the number of edges of $\kg$, i.e., $|V_L|=|E_G|$; \textit{(ii)} $|E_L| \propto \frac{1}{2} \sum_{v\in V_G} d_v^2 -|E_G|$, where $d_{v}$ denotes the degree of the node $v\in V_G$.

\begin{figure}[t]
	\includegraphics[width=\textwidth]{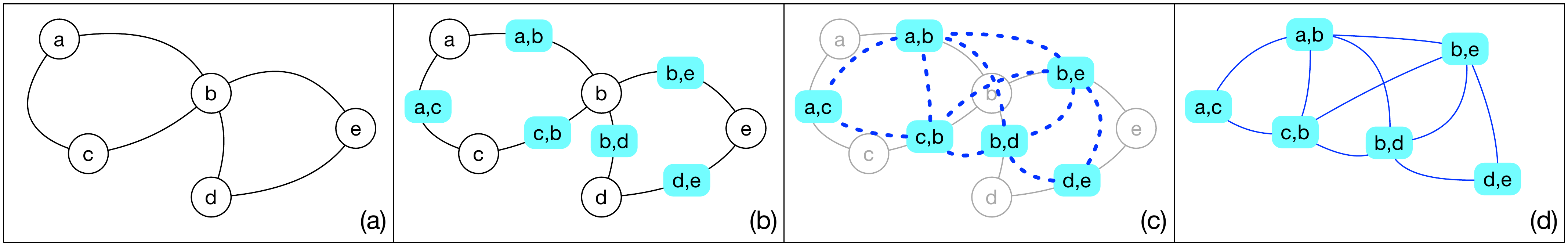}
	\caption{A undirected graph (a), its line graph (d); construction steps (b)-(c).}
	\label{fig:line_und_example}
\end{figure}

\begin{exmp}
	Each node of the line graph in Fig.~\ref{fig:line_und_example} (d) is labeled with the endpoints of the corresponding edge in the original graph Fig.~\ref{fig:line_und_example} (a). For instance, the node \emph{a,b} in Fig.~\ref{fig:line_und_example} (d) corresponds to the edge between the vertices \emph{a} and \emph{b} in Fig.~\ref{fig:line_und_example} (a). The node \emph{a,b} in Fig.~\ref{fig:line_und_example} (d) is adjacent to the node \emph{a,c} since the corresponding edges share the endpoint \emph{a} and to the vertices \emph{c,b}, \emph{b,d} and \emph{b,e} since the corresponding edges share the endpoint \emph{b} (see Fig.~\ref{fig:line_und_example} (a)).
	
\end{exmp}

The concept of line graph has been extended to other types of graphs, including multigraphs and directed graphs. On one hand, the extension to multigraphs adds a different node in the line graph from each edge of the original multigraph. On the other end, if $\kg$ is directed the corresponding line graph $\lkg$ will also be directed; in particular, its vertices are in one-to-one correspondence to the edges of $\kg$ and its edges represent {two-length directed paths} in $\kg$.
\begin{defn}\label{def:diline}
	Given a directed graph $G=(V_G,E_G)$, its line graph $\lkg=(V_L,E_L)$ is a directed graph such that: (i) each node of $\lkg$ represents an edge of $G$; (ii) two vertices of $\lkg$, say \emph{a,b} and \emph{c,d}; an edge connects \emph{a,b} and \emph{c,d} iff, \emph{b}=\emph{c}.
\end{defn}

\begin{figure}[!t]
	\includegraphics[width=\textwidth]{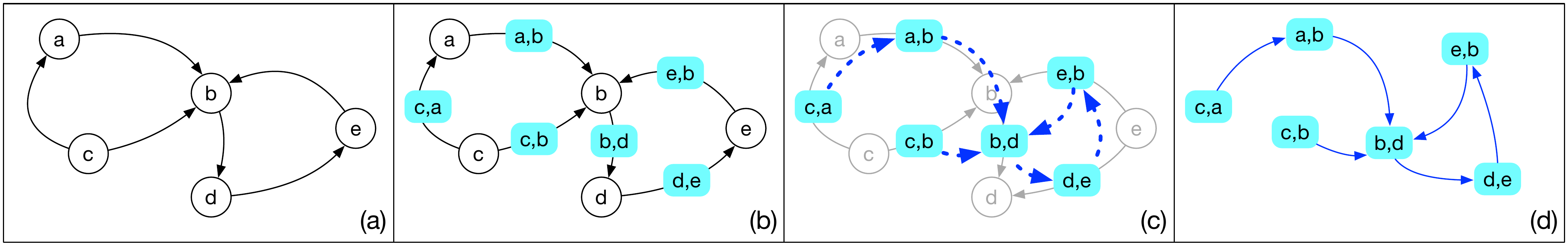}
	\caption{A directed graph (a), its line graph (d); construction steps (b)-(c).}
	\label{fig:line_dir_example}
\end{figure}

\begin{exmp}
In Fig.~\ref{fig:line_dir_example} (d),
each node of the directed line graph is labeled with the endpoints of the corresponding edge in the original directed graph. For instance, the node labeled \emph{a,b} in Fig.~\ref{fig:line_dir_example} (d) corresponds to the edge from the node \emph{a} to the node \emph{b} in Fig.~\ref{fig:line_dir_example} (a). The node \emph{a,b} in Fig.~\ref{fig:line_dir_example} (d) has an outgoing edge to the node \emph{b,d} since the corresponding edges share the intermediate endpoint \emph{b}, meaning that the edge from \emph{a,b} to \emph{b,d} encodes the directed two-length path from \emph{a}  to \emph{d} in $\kg$ (Fig.~\ref{fig:line_dir_example} (a)).
	
\end{exmp}

\begin{figure}[!t]
	\includegraphics[width=\textwidth]{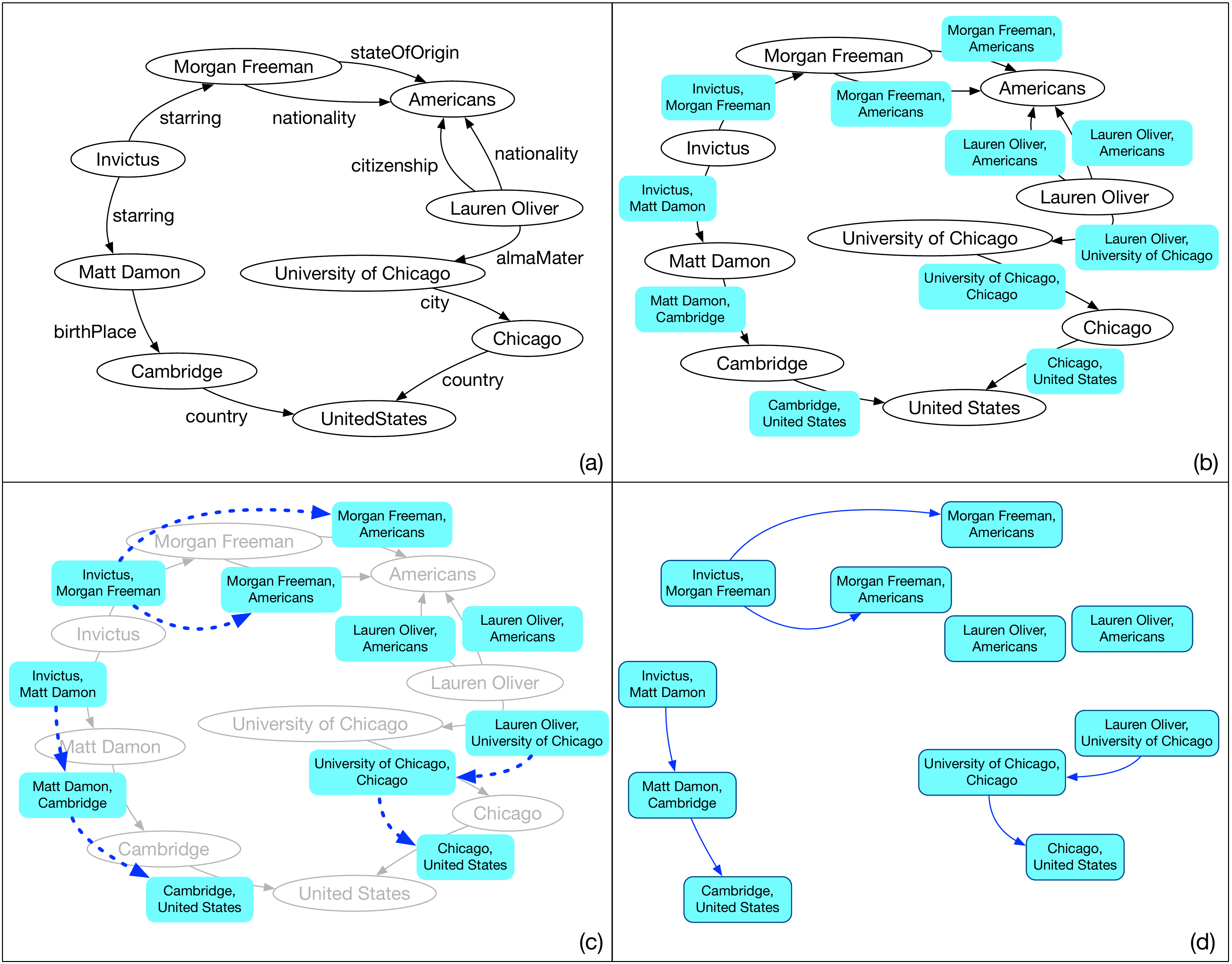}
	\caption{A knowledge graph (a), its triple line graph (d); construction steps (b)-(c).}
	\label{fig:diline_example}
\end{figure}
\section{The Triple Line Graph of a Knowledge Graph}
\label{sec:tripleLineGraph}
 The most natural way to apply the notion of line graph to knowledge graphs, that are directed labeled multigraphs, would be to apply Definition~\ref{def:diline}. However, this would lead to counter-intuitive behaviors. Consider the graph $\kg$  in Fig.~\ref{fig:diline_example} (a). Fig.~\ref{fig:diline_example} (b) shows in blue  nodes of the directed line graph $\lkg$, obtained by applying Definition~\ref{def:diline}, that are in one-to-one correspondence to the edges of $\kg$. Fig.~\ref{fig:diline_example} (c) shows how the directed edges are added to $\lkg$ to obtain the final directed line graph shown in Fig.~\ref{fig:diline_example} (d). At this point, three main issues arise. First, the standard definition associates to each node of the directed line graph the two endpoints of the corresponding edge; however, the edge labels in a knowledge graph give a semantic to the corresponding edge, which is completely lost if only the endpoints are considered.  Second, the edges to be added to the line graph are computed by considering their direction. This disregards the fact that edges in $\kg$ witness some semantic relation between their endpoints (i.e., entities) that can be interpreted bidirectionally. As an example, according to Definition~\ref{def:diline}, the two nodes \entity{(Lauren Oliver, Americans)} in $\lkg$ remain isolated since the corresponding edges do not belong to any two-length path in $\kg$ (see Fig.~\ref{fig:diline_example} (c)-(d)). However, consider the triple \entity{(\texttt{Lauren Oliver}, nationality, Americans)}. While the traversal of the  edge from \entity{Lauren Oliver} to \entity{Americans} serves the purpose of stating the relation \entity{nationality}, the traversal of the edge in the opposite direction states the relation \entity{is nationality of}. 
 Hence, \textit{in the case of knowledge graphs, two nodes of the line graph must be connected by an edge if they form a two-length path in the original knowledge graph no matter the edge direction, as the semantics of edges can be interpreted bidirectionally}. Finally, triples (via predicates) encode some semantic information, and the desideratum is to preserve this semantics when connecting two nodes (i.e., triples of $\kg$) in the line graph. Because of these issues, we introduce \textit{triple line graphs}, a novel notion of line graph suitable for \kgs.
 %%%%%%
%%%%%%

\begin{defn}
	Given a knowledge graph $G=(V_G,E_G,T_G)$, the associated triple line graph $\lkg=(V_L,E_L,w)$ is such that: (i) each node of $\lkg$ represents an edge of $G$; (ii) two vertices of $\lkg$, say $s_1,p_1,o_1$ and $s_2,p_2,o_2$, are adjacent if, and only if, $\{s_1,o_1\}\cap \{s_2,o_2\}\neq \emptyset$;  (iii) the function $w$ associates a weight in the range $[0,1]$ to each edge of $\lkg$.
\end{defn}

Given a Knowledge Graph $\kg=(V_G,E_G,T_G)$, its associated triple line graph $\lkg$ has $|T_G|$ nodes and each node of the original knowledge graph $\kg$, that is involved in $k$ triples, creates $k$ edges in the triple line graph.
\begin{figure}[!t]
	\includegraphics[width=\textwidth]{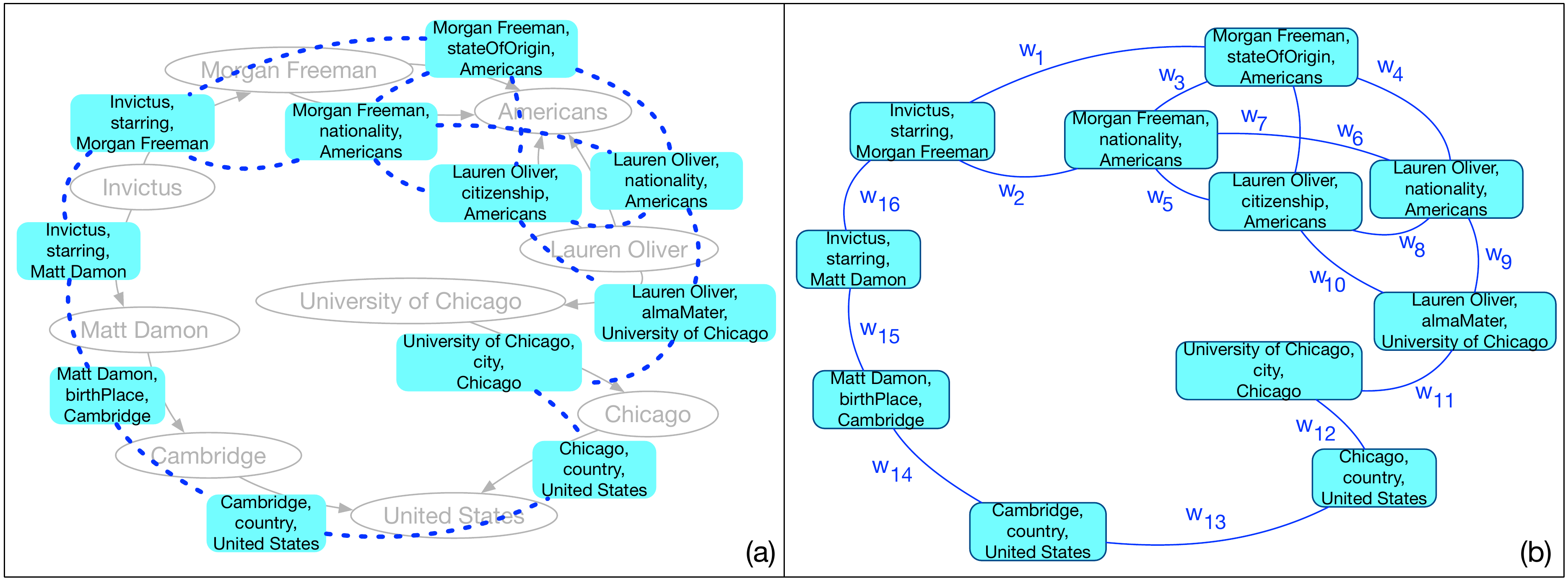}
	\caption{Subfigure (b) shows the triple line graph $\lkg$ associated to the knowledge graph $\kg$ in Fig.~\ref{fig:line_example} (a). Subfigure (a) shows the correspondence between the node of $\lkg$ and the triples of $\kg$.}
	\label{fig:line_example}
\end{figure}
\begin{exmp}
	Consider the graph $\kg$ reported in Figure~\ref{fig:diline_example} (a).  Fig.~\ref{fig:line_example} (b) shows the associated triple line graph $\lkg$, while in Figure~\ref{fig:line_example} (a) it is possible to identify the correspondence between the nodes of $\lkg$ and the triples in $\kg$. In particular, each node of the triple line graph is labeled with the subject predicate and object of the corresponding triple in $\kg$. For instance, the node labeled \entity{Invictus, starring, Matt Damon} in Fig.~\ref{fig:line_dir_example} (b) corresponds to the triple \entity{(Invictus, starring, Matt Damon)} in Fig.~\ref{fig:diline_example} (a). The nodes \entity{Invictus, starring, Matt Damon} and \entity{Invictus, starring, Morgan Freeman} in Fig.~\ref{fig:line_example} (b) are connected by an edge since the subject of the triple corresponding to the former node is also subject of the triple of the latter. Moreover, such an edge has associated the weight $w_{16}$ that should reflect the fact that the triples corresponding to the two endpoints share the same predicate \entity{starring}.
\end{exmp}
\subsection{Computing Triple Line Graphs}
\label{sec:line-graph-kg}
%%%%
We now describe an algorithm (outlined in Algorithm~\ref{alg:tlg}) to compute the triple line graph $\lkg$ of a knowledge graph $\kg$. This is at the core of \system\ for the computation of triple embeddings. After initializing the set of nodes and edges of $\lkg$  to the empty set (line 1), the algorithm iterates over the triples of the input $\kg$ and add a node to $\lkg$ for each visited triple (lines 2-3), thus inducing a bijection from the set of triples of $\kg$ to the set of nodes of $\lkg$ . Besides, if two triples share a node in $\kg$ then an edge will be added between the corresponding nodes of $\lkg$ (lines 5-14). In particular, the data structure $I(s)$ (line 6) keeps track, for each node $s$ of $\kg$, of the triples in which $s$ appears as subject or object (lines 7-8). Since such triples correspond to nodes of the triple line graph, by iterating over pairs of triples in $I(s)$ it is possible to add the desired edge between the corresponding nodes of $\lkg$ (lines 9-10). By scrutinizing this algorithm, one can see that there can be a large number of edges in the triple line graph. Therefore, we introduce a generic edge weighting mechanism (line 11). We are going to describe edge weighting mechanisms for both knowledge graphs and homogeneous graphs in Section \ref{sec:edge-weights}. We remark that in the  case of homogeneous graphs, the computation of line graphs is done by applying the standard algorithm.
%%
%\vspace{-.6cm}
%%
\begin{algorithm}[!h]
%	\scriptsize
	\SetKwInOut{Input}{Input}
	\SetKwInOut{Output}{Output}
	\Input{Knowledge Graph $G$}
	\Output{$\lkg$: the Triple Line Graph associated to $G$}
	\begin{algorithmic}[1]
		{
			\STATE $\lkg=\{\emptyset,\emptyset,\emptyset\}$
			\FORALL {$(s,p,o)$ in $\kg$}
			\STATE add the node $s,p,o$ to $\lkg$
			\ENDFOR
			\FORALL {$s\in \kg$} 
			\STATE $I(s)=\emptyset$
			\FORALL {$(s,p,o)$ (resp., $(o,p,s)$) in $G$}
			\STATE add $s,p,o$ (resp., $o,p,s$) to $I(s)$
			\FORALL {pair $n,n'$ in $I(s)$}
			\STATE add the edge $(n,n')$ to $\lkg$
			\STATE set $w(n,n')=\texttt{computeEdgeWeight}(n,n')$
			\ENDFOR
			\ENDFOR
			\ENDFOR
			\RETURN $L(\kg)$
		}
	\end{algorithmic}
	\caption{BuildTripleLineGraph ($\kg$)}
	\label{alg:tlg}
\end{algorithm}

By inspecting Algorithm~\ref{alg:tlg}, we observe that $\lkg$ can be computed in time $\mathcal{O}(|T|^2\times costWeight)$, where $costWeight$ is the cost of computing the weight between nodes in $\lkg$ (i.e., triples in $\kg$).
\section{\system: Learning Triple Embeddings}
\label{sec:triple-embeddings}
%%%
We now describe $\system$, which is the first approach for learning triple embeddings from knowledge graphs. \system\ includes four main phases: (i) \textit{building of the triple line graph} (outlined in Section \ref{sec:tripleLineGraph}); (ii) \textit{weighting of the triple line graph edges} (outlined in Section \ref{sec:edge-weights}); (iv)  \textit{computing walks} on the weighted triple line graph, described in Section \ref{sec:computing-walks}, and (v) \textit{computing embeddings} via the Skip-gram model, described in Section \ref{sec:computing-embeddings}.
\subsection{Triple Line Graph Edge Weighting}
\label{sec:edge-weights}
%%%%%%%
We have mentioned in Section \ref{sec:line-graph-kg} that the number of edges in the (triple) line graph can be large. This structure is much denser than that of the original graph and may significantly affect the performance of the walk generation strategy. To remedy this drawback, we introduce edge weighting functions (line 11 Algorithm \ref{alg:tlg}) for both knowledge and homogeneous graphs.
%%
%%%
%%%
\subsubsection{Relatedness Based Edge Weights for Knowledge Graphs.} The desideratum is to come up with a strategy to compute walks so that the neighborhood of a triple will include triples that are semantically related. We leverage a predicate relatedness measure~\cite{pirrobuilding}, which looks at the co-occurrence of each pair of predicates $p_i$ and $p_j$ in the set of triples $T$ and weights it by the predicate popularity~\cite{pirrobuilding}. Given two predicates $p_i$ and $p_j$, $Rel(p_i,p_j)$, $R(p_i,p_j)$ is obtained by computing the cosine between their respective vectors. In terms of edge weights, the more related predicates in the triples representing two nodes in the triple line graph are, the higher the weight of the edge between these nodes. Driving the walks via relatedness allows to capture both the graph topology in terms triple-to-triple relations  (i.e., edges in the triple line graph) and {semantic proximity} in terms of relatedness between predicates in triples.
%%%
%%%
\subsubsection{Centrality-based Edge Weights for Homogeneous Graphs.}
For the case of homogeneous graphs (e.g., social networks), we introduce an edge weighting  mechanism for $\lkg$ that relies on the notion of current-flow betweenness~\cite{BrandesF05}. This measure characterizes the importance of each node in $\kg$ in terms of the number of times it lies on a path between two other nodes (note that it extends the notion of betweenness centrality which focuses on shortest paths only). Therefore, a walker on $\kg$ would be directly affected by the current-flow betweenness of each node. To reflect the same behavior on the line graph $\lkg$, the edge weighting mechanism assigns a weight to the edge from $n_p=i,j$ to $n_q=j,k$ in $\lkg $ (representing a path from $i$ to $k$ passing through $j$ in $\kg$) proportional to the weighted mean among the current-flow betweenness centrality of the nodes $i$, $j$ and $k$ in the graph $G$. More formally, $w(n_p, n_q)$= $\alpha \cdot cb(i) + \beta\cdot  cb(j) + \gamma\cdot  cb(k)$ with $\alpha+\beta+\gamma=1$ and $cb(x)$ being the current-flow centrality of the node $x$, with $x \in \{i,j,k\}$.

\subsection{Computing Walks}
\label{sec:computing-walks}
\system\ leverages a language model approach to learn the final edge embeddings. As such, it requires a ``corpus" of  both nodes and sequences of nodes similarly to word embeddings techniques that require words and sequences of words (i.e., sentences). To obtain the corpus from the graph, we leverage truncated graph walks. The idea is to start from each node of $\lkg$ (representing an edge of the original graph) and provide a context for each of such node in terms of a sequence of other nodes.  Although walks have been used by previous approaches for both homogeneous (e.g., Deepwalk \cite{perozzi2014deepwalk}, node2vec \cite{grover2016node2vec}) and knowledge (e.g., metapath2vec, JUST \cite{hussein2018meta}) graphs, none of them has tackled the problem of computing edge embeddings. 

%%%
\subsection{Computing Embeddings}
\label{sec:computing-embeddings}
%%%
Once the ``corpus" (in terms of the set of walks $\mathcal{W}$) is available, the last step of the \system\ workflow is to compute the embeddings of the nodes of $\lkg$ that will correspond to the embeddings of the edges of the input graph $G$. The embedding we seek can be seen as a function $f:\mathcal{V}_L\rightarrow {R}^d$, which projects nodes of the weighted triple line graph $\lkg$ into a low dimensional vector space, where $d \ll |\mathcal{V}_L|$, so that neighboring nodes are in close proximity in the vector space. For every node $u \in \mathcal{V}_L$, $N(u)\subset \mathcal{V}_L$ is the set of neighbors, which is determined by the walks computed as described in Section \ref{sec:computing-walks}. The co-occurrence probability of two nodes $v_i$ and $v_{i+1}$ in a set of walks $\mathcal{W}$ is given by the softmax function using their vector embeddings $e_{v_i}$ and $e_{v_{i+1}}$:
\begin{equation}\label{eq:nextnode}
p((e_{v_i},e_{v_{i+1}})\in \mathcal{W}) =\sigma(e_{v_i}^Te_{v_{i+1}})
\end{equation}

\noindent
where $\sigma$ is the softmax function and $e_{v_i}^Te_{v_{i+1}}$ is the dot product of the vectors $e_{v_i}$ and $e_{v_{i+1}}$
As the computation of (\ref{eq:nextnode}) is computationally demanding~\cite{grover2016node2vec}, we use negative sampling to training the Skip-gram model. Negative sampling randomly selects nodes that do not appear together in a walk as negative examples, instead of considering all nodes in a graph. In particular, if a node $v_i$ appears in walk of another node $v_{i+1}$, then the vector embedding $e_{v_i}$ is closer to $e_{v_{i+1}}$ as compared to any other randomly chosen node. The probability that a node $v_i$ and a randomly chosen node $v_j$ \textit{do not} appear in a random walk starting from  $v_i$ is given by: 
\begin{equation}
p((e_{j},e_i)\not\in \mathcal{W}) = \sigma(-e_{v_i}^Te_{v_j})
\end{equation}
For any two nodes $v_i$ and $v_{i+1}$, the negative sampling objective of the Skip-gram model to be maximized is given by the following objective function:
\begin{equation}
\mathcal{O}(\theta)= \log\sigma(e_{v_i}^T e_{v_{i+1}}) + \sum_{j=1}^{k}\mathbb{E}_{v_j}[\log \sigma(-e_{v_i}^T e_{v_j})],
\end{equation}
where $\theta$ denotes the set of all parameters and $k$ is the number of negative samples. For the optimization of the objective function, we use the parallel asynchronous stochastic gradient descent algorithm~\cite{recht2011hogwild}.
%%%%
%%%%
%%%
%%%
\section{Experiments}
\label{sec:experiments}
%%%%
In this section, we report on an experimental evaluation of \system\ and comparison with related work. We describe the datasets and the experimental setting in Section \ref{sec:dataset-setting}. Then, we report experiments on knowledge graphs in Section \ref{sec:hete-graphs} and on homogeneous graphs in Section \ref{sec:homo-graphs}.
%%%%%%%%%%
%%%%%%%%%%
\subsection{Datasets and Experimental Setting}
\label{sec:dataset-setting}
We discuss experiments on both knowledge graphs and homogeneous graphs. For knowledge graphs, we used the following three real-world data sets. 
\begin{table}[]
	\centering
	\caption{Datasets used.}
	\begin{tabular}{|c|c|c|}
		\hline
		\textbf{Dataset}        & $|\textbf{V}|$ & $|\textbf{E}|$  \\ \hline
		DBLP           & 16K & 52K  \\ \hline
		Foursquare     & 30K & 83K  \\ \hline
		Yago           & 22K & 89K  \\ \hline
		Karate         & 34  &  78    \\ \hline
		Les Miserables & 77  & 254  \\ \hline
		USA Power Grid & 5K  & 6.6K \\ \hline
	\end{tabular}
\end{table}
In DBLP \cite{huang2017heterogeneous}, there are 4 different kinds of edges linking authors to papers, papers to venues, papers to papers, and papers to topics. In addition, authors are labeled with one among four labels (i.e., database, data mining, machine learning, and information retrieval). Foursquare  \cite{yang2016participatory} includes four different kinds of entities, that is, users, places, points of interests and timestamps. Each point of interest has also associated one among 10 labels. Yago, described in \cite{huang2017heterogeneous}, includes 5 types of edges interlinking movies to directors, actors and so forth. Moreover, each movie is assigned one or more among 5 available labels.

\subsection{Systems and Parameter Setting}
As \system\ is the first approach to tackle the problem of embedding edges in (knowledge) graphs, there is an intrinsic difficulty in finding competitors. Therefore, we considered some existing approaches to learn edge embeddings. \textit{For homogeneous graphs}, edge embeddings were computed by aggregating the embeddings of the endpoint nodes. As the average gave the best results, in what follows we only discuss this aggregation mechanism. \textit{For knowledge graphs}, we adopted the same methodology. Nevertheless, in this case, this strategy leads to counter-intuitive behaviors since nodes can be connected by more than one edge in general (see Section \ref{sec:tripleLineGraph}). This further underlines the need for specific edge embedding approaches like \system, which can handle these situations by directly embedding triples in \kgs. \system\ has been implemented in Python\footnote{Link omitted to preserve the anonymity of the review} using the networkx\footnote{\url{https://networkx.github.io}} and gensim\footnote{\url{https://radimrehurek.com/gensim}} libraries to handle graph computations and learn embeddings, respectively. For the evaluation, we considered the following baselines implemented in the StellarGraph library\footnote{\url{https://www.stellargraph.io}}.
\begin{itemize}
	\itemsep0em
	\item \textbf{DeepWalk}: it learns node embeddings via random walks fed into the Skip-gram model. As this approach was originally designed for homogeneous graphs, we applied it on knowledge graphs by ignoring node and edge types.
	\item \textbf{node2vec}: it improves upon DeepWalk in both the way random walks are generated and in the objective function optimization (it uses negative sampling). We set the parameters specific to this algorithm (i.e., $p$ and $q$) to the best values reported in \cite{grover2016node2vec}. For the same reason as Deepwalk, we apply node2vec by ignoring node and edge types.
	\item \textbf{Metapath2vec}: it has been defined to work on knowledge graphs. It takes as input one or more \textit{metapaths} (i.e., sequences of node types) to generate walks fed into a variant of the Skip-gram model. For the evaluation, we used the metapaths used in previous evaluations \cite{dong2017metapath2vec,hussein2018meta}. 
\end{itemize}
\begin{figure}[!t]
	%\hspace{1ex}
	\begin{tikzpicture}
	\begin{customlegend}[legend columns=7,legend style={align=left,draw=none,column sep=0ex},legend entries={\system,node2vec, Deepwalk, metapath2vec}]
	\addlegendimage{color=red, mark=ball}
	\addlegendimage{color=blue, mark=otimes}
	\addlegendimage{color=black, mark=star}
	\addlegendimage{color=yellow!60!red, mark=diamond}
	\end{customlegend}
	\end{tikzpicture}
	\begin{minipage}{0.3\textwidth}
		\begin{tikzpicture}[scale=0.45][font=\LARGE]
		\begin{axis}[
		xlabel={\% Labeled nodes from $\lkg$},
		ylabel={Micro-F1},
		title={DBLP},
		]
		\plotred file[x index=0, y index=0.2] {DBLP_triple2vec_micro.txt};
		
		\plotblue file[x index=0, y index=0.2] {DBLP_node2vec_micro.txt};
		
		\plotblack file[x index=0, y index=0.2] {DBLP_deepWalk_micro.txt};
		
		\plotyellow file[x index=0, y index=0.2] {DBLP_metapath2vec_micro.txt};
		
		\end{axis}
		\end{tikzpicture}
	\end{minipage}
	\begin{minipage}{0.3\textwidth}
		\begin{tikzpicture}[scale=0.45][font=\LARGE]
		\begin{axis}[
		xlabel={\% Labeled nodes from $\lkg$},
		title={Foursquare}
		]
		\plotred file[x index=0, y index=0.2] {Foursquare_triple2vec_micro.txt};
		
		\plotblue file[x index=0, y index=0.2] {Foursquare_node2vec_micro.txt};
		
		\plotblack file[x index=0, y index=0.2] {Foursquare_deepWalk_micro.txt};
		
		\plotyellow file[x index=0, y index=0.2]{Foursquare_metapath2vec_micro.txt};
		\end{axis}
		\end{tikzpicture}
	\end{minipage}
	\begin{minipage}{0.3\textwidth}
		\begin{tikzpicture}[scale=0.45][font=\LARGE]
		\begin{axis}[
		xlabel={\% Labeled nodes from $\lkg$},
		title={Yago},
		legend style={at={(-0.6,-0.3)},anchor=west}, 
		]
		\plotred file[x index=0, y index=0.2] {Movies_triple2vec_micro.txt};
		
		\plotblue file[x index=0, y index=0.2] {Movies_node2vec_micro.txt};
		
		\plotblack file[x index=0, y index=0.2] {Movies_deepWalk_micro.txt};
		
		\plotyellow file[x index=0, y index=0.2] {Movies_metapath2vec_micro.txt};
		\end{axis}
		\end{tikzpicture}
	\end{minipage}
	\begin{minipage}{0.3\textwidth}
		\begin{tikzpicture}[scale=0.45][font=\LARGE]
		\begin{axis}[
		xlabel={\% Labeled nodes from $\lkg$},
		ylabel={Macro-F1},
		title={DBLP}
		]
		\plotred file[x index=0, y index=0.2] {DBLP_triple2vec_macro.txt};
		
		\plotblue file[x index=0, y index=0.2] {DBLP_node2vec_macro.txt};
		
		\plotblack file[x index=0, y index=0.2] {DBLP_deepWalk_macro.txt};
		
		\plotyellow file[x index=0, y index=0.2] {DBLP_metapath2vec_macro.txt};
		
		\end{axis}
		\end{tikzpicture}
	\end{minipage}
	\hspace{0.4cm}
	\begin{minipage}{0.3\textwidth}
		\begin{tikzpicture}[scale=0.45][font=\LARGE]
		\begin{axis}[
		xlabel={\% Labeled nodes from $\lkg$},
		title={Foursquare}
		]
		\plotred file[x index=0, y index=0.2] {Foursquare_triple2vec_macro.txt};
		
		\plotblue file[x index=0, y index=0.2] {Foursquare_node2vec_macro.txt};
		
		\plotblack file[x index=0, y index=0.2] {Foursquare_deepWalk_macro.txt};
		
		\plotyellow file[x index=0, y index=0.2] {Foursquare_metapath2vec_macro.txt};
		
		\end{axis}
		\end{tikzpicture}
	\end{minipage}
	\hspace{0.4cm}
	\begin{minipage}{0.3\textwidth}
		\begin{tikzpicture}[scale=0.45][font=\LARGE]
		\begin{axis}[
		legend style={at={(-0.3,-0.3)},anchor=west},
		legend columns=-1,
		xlabel={\% Labeled nodes from $\lkg$},
		title={Yago}
		]
		\plotred file[x index=0, y index=0.2] {Movies_triple2vec_macro.txt};
		
		\plotblue file[x index=0, y index=0.2] {Movies_node2vec_macro.txt};
		
		\plotblack file[x index=0, y index=0.2] {Movies_deepWalk_macro.txt};
		
		\plotyellow file[x index=0, y index=0.2] {Movies_metapath2vec_macro.txt};
		%\addlegendentry{Metapath2vec};
		%
		\end{axis}
		\end{tikzpicture}
	\end{minipage}
	\caption{Triple classification results in terms of micro and macro F1.}
	\label{fig:classification-kg}
\end{figure}
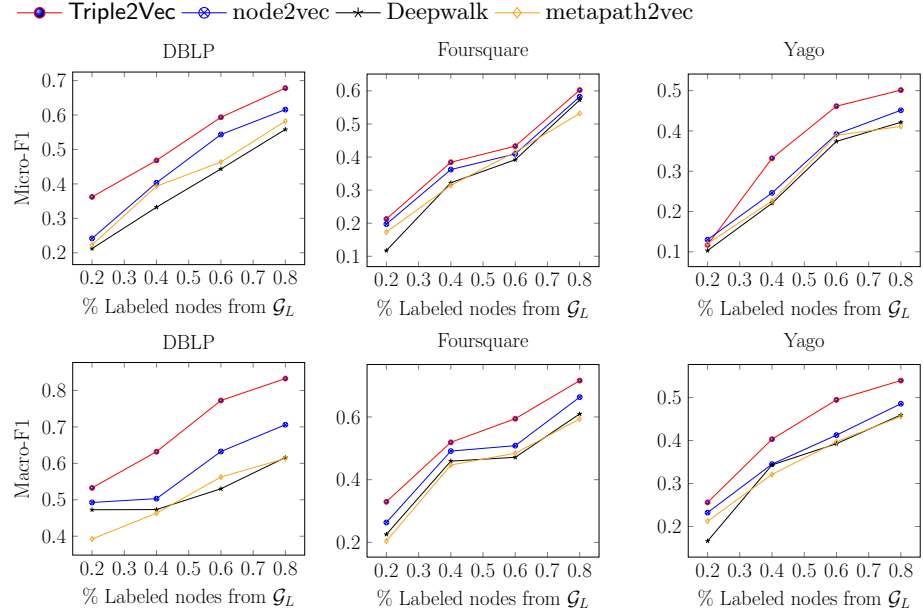
%%%%%%%%%
%%%%%%%%%

For sake of space, in what follows, we only report the best results obtained by setting the parameters as follows: the number of walks per node $n$=$10$, maximum walk length $L=100$, the window size (necessary for the notion of context in the Skip-gram model) $w=10$.
Moreover, we used $d$=128 as a dimension of the node embeddings for DBLP, Foursquare and Yago and $d$=32 for the other datasets. The number of negative samples $\Gamma$ is set to 10 for all methods in all experiments. All results reported are the average of 10 runs.
%%%%%
%%%%%

\begin{figure}%
	\centering
	\subfloat[Triple2Vec]{{\includegraphics[width=5cm]{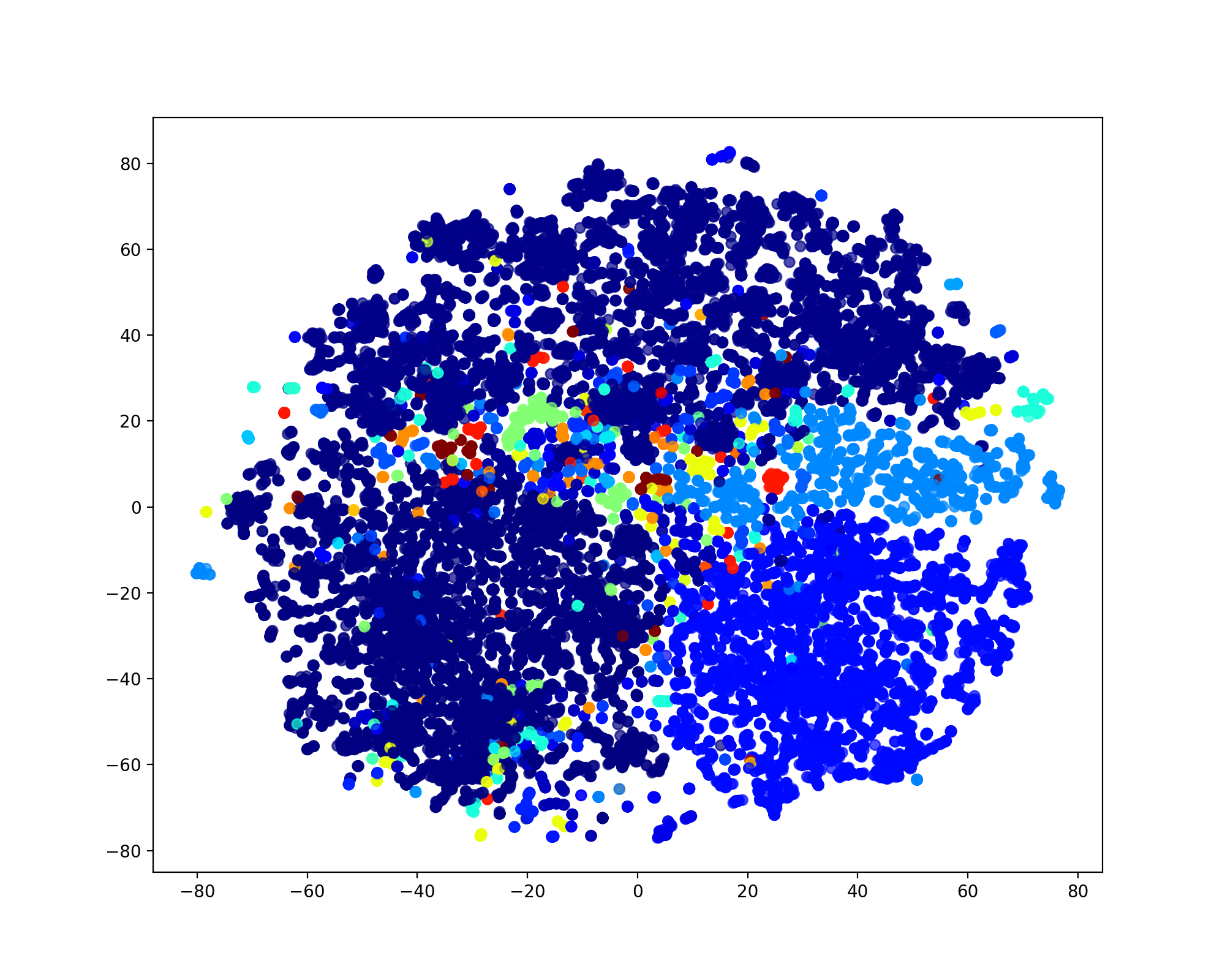} }}%
	\qquad
	\subfloat[metapath2vec]{{\includegraphics[width=5cm]{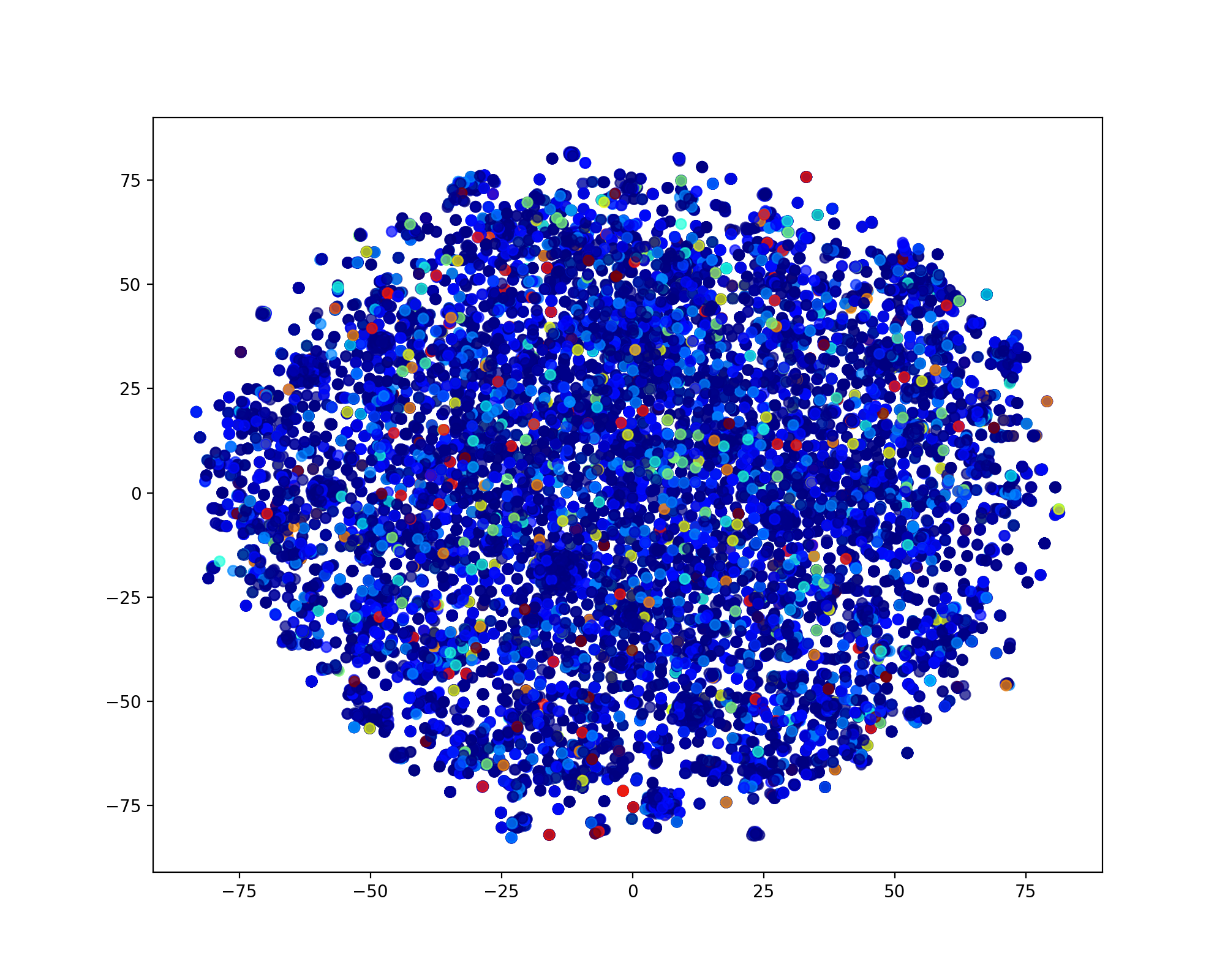} }}%
	\caption{Triple embedding visualization of the DBLP dataset.}%
	\label{fig:emb-DBLP}%
\end{figure}
\subsection{Evaluation on Knowledge Graphs}
\label{sec:hete-graphs}
%%%%%%%
%%%%%%
To the best of our knowledge \system\ is the first approach to specifically tackle the problem of embeddings triples from heterogeneous graphs like knowledge graphs. As such, there was no benchmark available. In order to construct a benchmark for the evaluation, we labeled the triples of the datasets by propagating the available labels. Specifically, for DBLP the 4 author node labels available have been propagated to paper nodes by following authorship links; then, from papers nodes to topic and venue nodes. For Yago, movie nodes were labeled with 12 labels that have been propagated to actors, musicians and directors by following \entity{actedIn}, \entity{wroteMusicFor} and \entity{directed} edges, respectively. Finally, points of interest for FourSquare were labeled with 10 labels that have been propagated to places, users and timestamps by following \entity{locate}, \entity{perform} and \entity{happendAt} edges, respectively. At the end of this label propagation step, each node of $\kg$ is labeled with a subset of the initial labels. To propagate these labels to the nodes of $\lkg$, we considered the union of the sets of labels associated with the node endpoints of the corresponding triples in $\kg$ represented by the nodes of $\lkg$. Finally, to each different subset of labels, we assigned a different label. We now report on the evaluation on two different tasks.
%%%%%%%%%
%%%%%%%%%

\noindent
\textbf{Triple Classification.} In order to carry out this task, we trained a one-vs-rest Logistic regression model, giving as input the triple embeddings along with their labels (the labels of the node of $\lkg$). Then, we compute the Micro-F1 and Macro-F1 scores by varying the percentage of training data. Results are reported in Fig. \ref{fig:classification-kg}. We observe that \system\ consistently outperforms the baseline. This is especially true in the DBLP and Yago datasets. We also note that metapath2vec performs worse than node2vec and deepwalk, despite the fact that the former has been proposed to work on knowledge graphs. This may be explained by the fact that the metapaths used in the experiments, while being able to capture node embeddings, fail short in capturing edge (triple) embeddings.

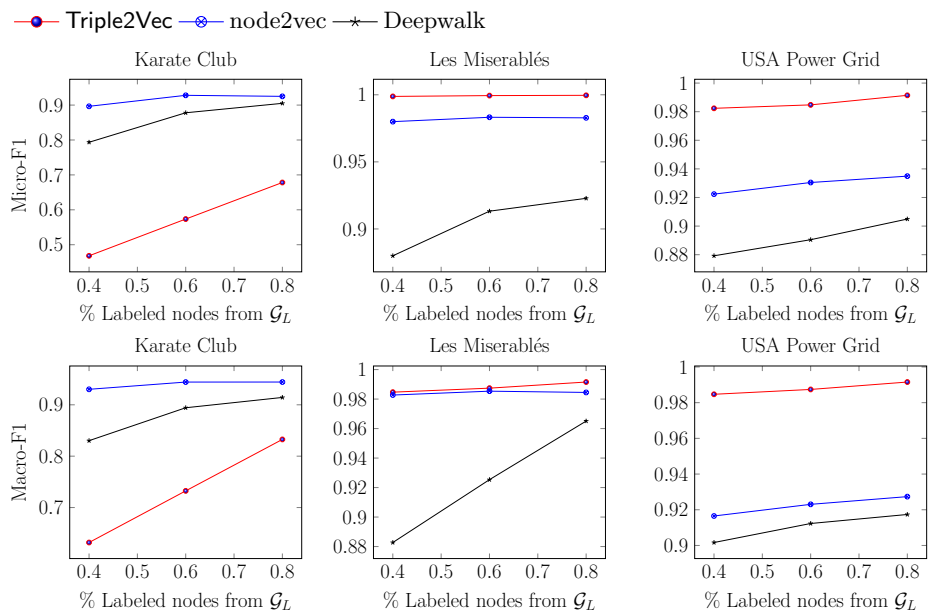
\begin{figure}[h!]
	%\hspace{1ex}
	\begin{tikzpicture}
	\begin{customlegend}[legend columns=7,legend style={align=left,draw=none,column sep=0ex},legend entries={\system,node2vec, Deepwalk}]
	\addlegendimage{color=red, mark=ball}
	\addlegendimage{color=blue, mark=otimes}
	\addlegendimage{color=black, mark=star}
	\end{customlegend}
	\end{tikzpicture}
	
	\begin{minipage}{0.3\textwidth}
		\begin{tikzpicture}[scale=0.45][font=\LARGE]
		\begin{axis}[
		xlabel={\% Labeled nodes from $\lkg$},
		ylabel={Micro-F1},
		title={Karate Club},
		]
		\plotred file[x index=0, y index=0.2] {karate_triple2vec_micro.txt};
		
		\plotblue file[x index=0, y index=0.2] {karate_node2vec_micro.txt};
		
		\plotblack file[x index=0, y index=0.2] {karate_deepWalk_micro.txt};
		\end{axis}
		\end{tikzpicture}
	\end{minipage}
	\begin{minipage}{0.3\textwidth}
		\begin{tikzpicture}[scale=0.45][font=\LARGE]
		\begin{axis}[
		xlabel={\% Labeled nodes from $\lkg$},
		title={Les Miserabl\'es}
		]
		\plotred file[x index=0, y index=0.2] {lesmis_triple2vec_micro.txt};
		
		\plotblue file[x index=0, y index=0.2] {lesmis_node2vec_micro.txt};
		
		\plotblack file[x index=0, y index=0.2] {lesmis_deepWalk_micro.txt};
		\end{axis}
		\end{tikzpicture}
	\end{minipage}
	\begin{minipage}{0.3\textwidth}
		\begin{tikzpicture}[scale=0.45][font=\LARGE]
		\begin{axis}[
		xlabel={\% Labeled nodes from $\lkg$},
		title={USA Power Grid},
		legend style={at={(-0.6,-0.3)},anchor=west}, 
		]
		\plotred file[x index=0, y index=0.2] {power_triple2vec_micro.txt};
		
		\plotblue file[x index=0, y index=0.2] {power_node2vec_micro.txt};
		
		\plotblack file[x index=0, y index=0.2] {power_deepWalk_micro.txt};
		\end{axis}
		\end{tikzpicture}
	\end{minipage}
	\begin{minipage}{.3\textwidth}
		\begin{tikzpicture}[scale=0.45][font=\LARGE]
		\begin{axis}[
		xlabel={\% Labeled nodes from $\lkg$},
		ylabel={Macro-F1},
		title={Karate Club}
		]
		\plotred file[x index=0, y index=0.2] {karate_triple2vec_macro.txt};
		
		\plotblue file[x index=0, y index=0.2] {karate_node2vec_macro.txt};
		
		\plotblack file[x index=0, y index=0.2] {karate_deepWalk_macro.txt};
		
		\end{axis}
		\end{tikzpicture}
	\end{minipage}
	\hspace{0.4cm}
	\begin{minipage}{0.3\textwidth}
		\begin{tikzpicture}[scale=0.45][font=\LARGE]
		\begin{axis}[
		xlabel={\% Labeled nodes from $\lkg$},
		title={Les Miserabl\'es}
		]
		\plotred file[x index=0, y index=0.2] {lesmis_triple2vec_macro.txt};
		
		\plotblue file[x index=0, y index=0.2] {lesmis_node2vec_macro.txt};
		
		\plotblack file[x index=0, y index=0.2] {lesmis_deepWalk_macro.txt};
		\end{axis}
		\end{tikzpicture}
	\end{minipage}
	\hspace{0.4cm}
	\begin{minipage}{0.3\textwidth}
		\begin{tikzpicture}[scale=0.45][font=\LARGE]
		\begin{axis}[
		legend style={at={(-0.3,-0.3)},anchor=west},
		legend columns=-1,
		xlabel={\% Labeled nodes from $\lkg$},
		title={USA Power Grid}
		]
		\plotred file[x index=0, y index=0.2] {power_triple2vec_macro.txt};
		
		\plotblue file[x index=0, y index=0.2] {power_node2vec_macro.txt};
		
		\plotblack file[x index=0, y index=0.2] {power_deepWalk_macro.txt};
		%\addlegendentry{Metapath2vec};
		%
		\end{axis}
		\end{tikzpicture}
	\end{minipage}
	\caption{Edge classification results in terms of Micro and Macro F1.}
	\label{fig:classification-homo}
\end{figure}

\noindent
\textbf{Triple Clustering and visualization.} To have a better account of how triple embeddings are placed in the embedding space, we used t-SNE \cite{maaten2008visualizing} to obtain a 2-d representation of the triple embeddings (originally including $d$ dimensions) obtained from  \system\ and metapath2vec, which is the only one among the competitors, designed for knowledge graphs. Results are shown in Fig. \ref{fig:emb-DBLP} on DBLP. We observed similar trends in the other datasets. We note that while \system\ is able to clearly identify groups of triples (i.e., triples labeled with the same labels), metapath2vec offers a less clear perspective. We can explain this behavior with the fact that \system\ defines a specific strategy for triple embeddings based on the notion of semantic proximity, while triple embeddings metapath2vec have been obtained from the endpoint nodes.
%%%
%%%

%
%
\subsection{Evaluation on Homogeneous Graphs}
\label{sec:homo-graphs}
For the case of homogeneous  graphs, we compared \system\ with Deepwalk and node2vec only. This is because metapath2vec requires metapaths as input that are not available in homogeneous graphs. The evaluation was carried as follows. For each graph, we first found node communities by using a modularity-based algorithm that does not return overlapping communities \cite{newman2006modularity}. Then, for each community, returned as a set of nodes, we identify the set of intra-community edges and labeled each of such edges with the id of the community it belongs to.

\noindent
\textbf{Edge Classification.} As for the case of knowledge graphs, we trained a one-vs-rest Logistic regression model, giving as input the edge embeddings and the labels (the community they belong to). Results are reported in Fig. \ref{fig:classification-homo}. We can notice that even in this case \system\ performs better than the baselines. In particular, the difference with the baselines becomes clearer when moving to larger networks (from left to right in Fig. \ref{fig:classification-homo}).
%%
%%

%%
%%%
\subsubsection{Edge Clustering.} 
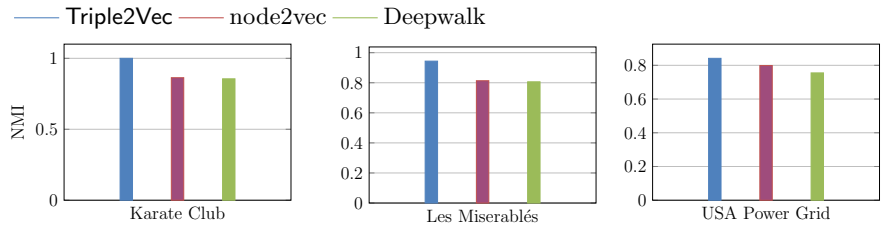
\begin{figure}[!t]
	\begin{tikzpicture}
	\begin{customlegend}[legend columns=7,legend style={align=left,draw=none,column sep=0ex},legend entries={\system, node2vec, Deepwalk}]
	\addlegendimage{bblue,fill=bblue,mark=none}
	\addlegendimage{rred,fill=ppurple,mark=none}   
	\addlegendimage{ggreen,fill=ggreen,mark=none}
	\end{customlegend}
	\end{tikzpicture}
	
	\begin{minipage}{0.3\textwidth}
		\begin{tikzpicture}[scale=0.47][font=\Large]
		\begin{axis}[
		width  = 8cm, %0.95*\textwidth,
		height = 6cm,
		xlabel = {Karate Club},
		%major x tick style = transparent,
		ybar=2*\pgflinewidth,
		bar width=10pt,
		ymajorgrids = true,
		ylabel = {NMI},
		symbolic x coords={T,N,D}, 
		scaled y ticks = false,
		enlarge x limits=1,
		ymin=0,
		%ticks=none,
		xmajorticks=false,
		legend pos=outer north east]
		]
		\addplot[style={bblue,fill=bblue,mark=none}] coordinates {(T,1.0)};
		\addplot[style={rred,fill=ppurple,mark=none}] coordinates {(N,0.8634)};
		\addplot[style={ggreen,fill=ggreen,mark=none}] coordinates {(D,0.8556)};
		\end{axis}
		\end{tikzpicture}
	\end{minipage}
	\hspace{0.5cm}
	\begin{minipage}{0.3\textwidth}
		\begin{tikzpicture}[scale=0.47][font=\Large]
		\begin{axis}[
		width  = 8cm, %0.95*\textwidth,
		height = 6cm,
		xlabel = {Les Miserabl\'es},
		%major x tick style = transparent,
		ybar=2*\pgflinewidth,
		bar width=10pt,
		ymajorgrids = true,
		symbolic x coords={\system,node2vec,DeepWalk}, 
		scaled y ticks = false,
		enlarge x limits=1,
		ymin=0,
		xmajorticks=false,
		%x tick label style={font=,text width=.7cm,rotate=65},
		legend pos=outer north east]
		]
		\addplot[style={bblue,fill=bblue,mark=none}] coordinates {(\system,0.9438)};
		\addplot[style={rred,fill=ppurple,mark=none}] coordinates {(node2vec,0.8134)};
		\addplot[style={ggreen,fill=ggreen,mark=none}] coordinates {(DeepWalk,0.8063)};
		%\legend{\system,node2vec,DeepWalk,Metapath2vec,JUST}
		\end{axis}
		\end{tikzpicture}		
	\end{minipage}
	%\hspace{0.5cm}
	\begin{minipage}{0.3\textwidth}
		\begin{tikzpicture}[scale=0.47][font=\Large]
		\begin{axis}[
		width  = 8cm, %0.95*\textwidth,
		height = 6cm,
		xlabel = {USA Power Grid},
		%major x tick style = transparent,
		ybar=2*\pgflinewidth,
		bar width=10pt,
		ymajorgrids = true,
		symbolic x coords={\system,node2vec,DeepWalk}, 
		scaled y ticks = false,
		enlarge x limits=1,
		ymin=0,
		y tick label style={
			/pgf/number format/fixed
		},
		xmajorticks=false,
		legend cell align=left,
		legend style={
			at={(0.5,-0.2)},
			anchor=north,legend columns=-1,
			column sep=1ex
		}
		]
		\addplot[style={bblue,fill=bblue,mark=none}] coordinates {(\system,0.8415)};
		\addplot[style={rred,fill=ppurple,mark=none}] coordinates {(node2vec,0.7986)};
		\addplot[style={ggreen,fill=ggreen,mark=none}] coordinates {(DeepWalk,0.7551)};
		%	\legend{\system,node2vec,DeepWalk,Metapath2vec,JUST}
		\end{axis}
		\end{tikzpicture}
	\end{minipage}
	\caption{Clustering results.}
	\label{fig:clustering}
\end{figure}

We also evaluate the performance of the systems on a clustering task. In particular, we ran the K-means algorithm giving the edge embeddings as input. Then, we compute the Normalized Mutual Information (NMI). Results are reported in Fig. \ref{fig:clustering}. Again, \system\ performs better than the competitors. We want to mention the fact that we are reporting the best performance of node2vec and Deepwalk in terms of the aggregation mechanism on node embeddings. On the contrary, \system\ does not requires any aggregation being focused on directly learning edge embeddings.

%%%
%%%
\section{Related Work}
\label{sec:related-work}
%%%%
 There is a vast body of related research about graph embedding techniques for both homogeneous \cite{cai2018comprehensive} and knowledge graphs \cite{wang2017knowledge}. In what follows we focus our attention on node embeddings that are the closest to our approach.
Early work  on node  embeddings has focused on homogeneous graphs, that is, graphs having one type of node and edge only~\cite{grover2016node2vec,perozzi2014deepwalk,tang2015line}. DeepWalk~\cite{perozzi2014deepwalk}, node2vec~\cite{grover2016node2vec} and LINE~\cite{tang2015line} are inspired by word2vec~\cite{mikolov2013distributed}, which has been proposed to learn the vector representations of words that appear in a text corpus. These techniques sample a set of random walks (and, in particular, they differ in how they sample the walks) in the original graph that is fed to a Skip-gram model to generate the vector representation of nodes, so that two nodes that frequently co-occurr in a randomly sampled path will have similar embeddings. DeepWalk~\cite{perozzi2014deepwalk} generates short truncated random walks by uniformly sampling the starting node and the additional nodes from the neighbors of the last node visited.
node2vec~\cite{grover2016node2vec} generates biased random walks by using two parameters to control how fast the walk explores and leaves the neighborhood of a node.
LINE~\cite{tang2015line} guides the generation of random walks by using 1-hop and 2-hop neighborhoods of nodes as such it learns two different latent representations of nodes. 

Another strand of research has focused on heterogeneous graphs, where nodes and edges can have different types~\cite{dong2017metapath2vec,fu2017hin2vec,hussein2018meta,ristoski2016rdf2vec}. Here, the random walk generation for the Skip-gram model has been adapted to consider nodes and edge types. 
RDF2Vec~\cite{ristoski2016rdf2vec} focuses on computing node embeddings by using the continuous bag of words or a Skip-gram model. It computes two kinds of walks: subtrees up to a fixed depth $k$ and breadth-first search walks(by uniformly sampling the nodes on the walks among the neighbors). metapath2vec~\cite{dong2017metapath2vec} uses metapaths to guide the generation of walks, but it also proposes to use heterogeneous negative samples in the Skip-gram model for learning latent vectors of nodes. Hin2vec~\cite{fu2017hin2vec} is an evolution of metapath2vec, which considers multiple metapaths. JUST~\cite{hussein2018meta} provides a sampling strategy that balances both the presence of homogeneous, heterogeneous edges and the node distribution over different domains (i.e., node types) in the generated walks. Our approach is also different from TransE \cite{bordes2013translating} and its variants, the goal of which is to learn knowledge graph embeddings to perform link prediction by providing both positive and negative input facts.

To the best of our knowledge, \system\ is the first approach that focuses on embedding triples in knowledge graphs. We already mentioned that approaches line node2vec have proposed ways to learn embeddings for edges as some combination of embeddings of the edge endpoints (e.g., Hadamard product, average). Despite the fact that this approach is inherently sub-optimal, when applied to knowledge graphs it will lead to counter-intuitive behaviors. Indeed, all the triples involving the same pair of nodes will be given the same vector representation. \system\ specifically focuses on learning triple embeddings guided by semantic proximity, which takes into account the semantics of edges. Finally, we also devised a novel edge weighting mechanism for homogeneous graphs.

\section{Concluding Remarks and Future Work}
\label{sec:conclusions}
%%%%
We introduced the novel task of learning edges from (knowledge) graphs. While for homogeneous graphs, there have been some sub-optimal proposals \cite{grover2016node2vec}, for knowledge graphs, the problem of learning triple embeddings, was never explored. We presented an elegant solution, which builds upon the notion line graph and extends it to knowledge graphs. We introduced semantic proximity as a way to place together triples expressed with related predicates close in the embedding space. We have also considered the case of homogeneous graphs, where node proximity is preserved via the notion of current-flow betweenness. The assembling of these novel ideas in the \system\ system can pave the way to a novel class of applications based on triple embeddings. We have discussed in the experiments tasks related to triple classification and visualization, for knowledge graphs, and edge classification and community detection for homogeneous graphs. There is still a lot to be explored, from novel ways of imposing triple proximity (e.g., via constraints \cite{minervini2017regularizing}) to novel applications like fact-checking.

\bibliographystyle{plain}
\bibliography{biblio-embeddings}

\begin{thebibliography}{10}

\bibitem{bordes2013translating}
A.~Bordes, N.~Usunier, A.~Garcia-Duran, J.~Weston, and O.~Yakhnenko.
\newblock Translating embeddings for modeling multi-relational data.
\newblock In {\em NIPS}, 2013.

\bibitem{BrandesF05}
U.~Brandes and D.~Fleischer.
\newblock Centrality measures based on current flow.
\newblock In {\em STACS}, pages 533--544.

\bibitem{cai2018comprehensive}
H.~Cai, V.~W Zheng, and K.~C.-C. Chang.
\newblock A comprehensive survey of graph embedding: Problems, techniques, and
  applications.
\newblock {\em TKDE}, 30(9):1616--1637, 2018.

\bibitem{dong2017metapath2vec}
X.~Dong, N.~V Chawla, and A.~Swami.
\newblock metapath2vec: Scalable representation learning for heterogeneous
  networks.
\newblock In {\em CIKM}, pages 135--144, 2017.

\bibitem{fu2017hin2vec}
T.-Y. Fu, W.-C. Lee, and Z.~Lei.
\newblock Hin2vec: Explore meta-paths in heterogeneous information networks for
  representation learning.
\newblock In {\em CIKM}, pages 1797--1806, 2017.

\bibitem{grover2016node2vec}
A.~Grover and J.~Leskovec.
\newblock node2vec: Scalable feature learning for networks.
\newblock In {\em KDD}, pages 855--864, 2016.

\bibitem{huang2017heterogeneous}
Z.~Huang and N.~Mamoulis.
\newblock Heterogeneous information network embedding for meta path based
  proximity.
\newblock {\em arXiv preprint arXiv:1701.05291}, 2017.

\bibitem{hussein2018meta}
R.~Hussein, D.~Yang, and P.~Cudr{\'e}-Mauroux.
\newblock Are meta-paths necessary?: Revisiting heterogeneous graph embeddings.
\newblock In {\em CIKM}, pages 437--446. ACM, 2018.

\bibitem{maaten2008visualizing}
L.~van~der Maaten and G.~Hinton.
\newblock Visualizing data using t-sne.
\newblock {\em Journal of machine learning research}, 9:2579--2605, 2008.

\bibitem{mikolov2013distributed}
T.~Mikolov, I.~Sutskever, K.~Chen, G.~S Corrado, and J.~Dean.
\newblock Distributed representations of words and phrases and their
  compositionality.
\newblock In {\em NIPS}, pages 3111--3119, 2013.

\bibitem{minervini2017regularizing}
P.~Minervini, L.~Costabello, E.~Mu{\~n}oz, V.~Nov{\'a}{\v{c}}ek, and P.-Y.
  Vandenbussche.
\newblock Regularizing knowledge graph embeddings via equivalence and inversion
  axioms.
\newblock In {\em PKDD}, pages 668--683, 2017.

\bibitem{newman2006modularity}
M.~EJ Newman.
\newblock Modularity and community structure in networks.
\newblock {\em Proceedings of the national academy of sciences},
  103(23):8577--8582, 2006.

\bibitem{perozzi2014deepwalk}
B.~Perozzi, R.~Al-Rfou, and S.~Skiena.
\newblock Deepwalk: Online learning of social representations.
\newblock In {\em KDD}, pages 701--710, 2014.

\bibitem{pirrobuilding}
G.~Pirr{\`o}.
\newblock Building relatedness explanations from knowledge graphs.
\newblock {\em Semantic Web}, (Preprint):1--28.

\bibitem{recht2011hogwild}
B.~Recht, C.~Re, S.~Wright, and F.~Niu.
\newblock Hogwild: A lock-free approach to parallelizing stochastic gradient
  descent.
\newblock In {\em NIPS}, pages 693--701, 2011.

\bibitem{ristoski2016rdf2vec}
P.~Ristoski and H.~Paulheim.
\newblock Rdf2vec: Rdf graph embeddings for data mining.
\newblock In {\em ISWC}, pages 498--514, 2016.

\bibitem{tang2015line}
J.~Tang, M.~Qu, M.~Wang, M.~Zhang, J.~Yan, and Q.~Mei.
\newblock Line: Large-scale information network embedding.
\newblock In {\em WWW}, pages 1067--1077, 2015.

\bibitem{wang2017knowledge}
Q.~Wang, Z.~Mao, B.~Wang, and L.~Guo.
\newblock Knowledge graph embedding: A survey of approaches and applications.
\newblock {\em TKDE}, 29(12):2724--2743, 2017.

\bibitem{west1996introduction}
D.~Brent West et~al.
\newblock {\em Introduction to graph theory}, volume~2.
\newblock Prentice Hall, 1996.

\bibitem{yang2016participatory}
D.~Yang, D.~Zhang, and B.~Qu.
\newblock Participatory cultural mapping based on collective behavior data in
  location-based social networks.
\newblock {\em ACM TIST}, 7(3):30, 2016.

\end{thebibliography}
\end{document}